\documentclass[nohyperref]{article}

\usepackage[pagebackref,breaklinks,colorlinks,citecolor=cvprblue]{hyperref}
\usepackage{graphicx}
\usepackage{amsmath}
\usepackage{amsthm}
\usepackage{amssymb}
\usepackage{subcaption}
\usepackage{booktabs}
\usepackage{xcolor}
\usepackage{multirow}
\newtheorem{theorem}{\textbf{Theorem}}

\newtheorem{Remark}{\textbf{Remark}}

\usepackage{booktabs}

\usepackage[accepted]{icml2022}

\icmltitlerunning{Submission and Formatting Instructions for ICML 2022}

\begin{document}

\twocolumn[
\icmltitle{RankMatch: A Novel Approach to Semi-Supervised Label Distribution Learning Leveraging Inter-label Correlations}



\icmlsetsymbol{equal}{*}

\begin{icmlauthorlist}
	\icmlauthor{Zhiqiang Kou}{equal,seu}
	\icmlauthor{Yucheng Xie}{equal,seu}
	\icmlauthor{Jing Wang}{seu}
	\icmlauthor{Boyu shi}{seu}
	\icmlauthor{Yuheng Jia}{seu}
	\icmlauthor{Xin Geng}{seu}
\end{icmlauthorlist}

\icmlaffiliation{seu}{MOE Key Laboratory of Computer Network and Information Integration, School of Computer Science and Engineering, Southeast University, Nanjing, China}

\icmlcorrespondingauthor{Xin Geng}{xgeng@seu.edu.com}

\icmlkeywords{Machine Learning, ICML}

\vskip 0.3in
]



\printAffiliationsAndNotice{\icmlEqualContribution} 

\begin{abstract}
	This paper introduces RankMatch, an innovative approach for Semi-Supervised Label Distribution Learning (SSLDL). Addressing the challenge of limited labeled data, RankMatch effectively utilizes a small number of labeled examples in conjunction with a larger quantity of unlabeled data, reducing the need for extensive manual labeling in Deep Neural Network (DNN) applications.  Specifically, RankMatch introduces an ensemble learning-inspired averaging strategy that creates a pseudo-label distribution from multiple weakly augmented images. This not only stabilizes predictions but also enhances the model's robustness.  Beyond this, RankMatch integrates a pairwise relevance ranking (PRR) loss, capturing the complex inter-label correlations and ensuring that the predicted label distributions align with the ground truth.
	We establish a theoretical generalization bound for RankMatch, and through extensive experiments, demonstrate its superiority in performance against existing SSLDL methods. 
\end{abstract}
\section{Introduction}
\label{sec:intro}

\begin{figure}[!h]
	\centering
	\includegraphics[width=0.5\textwidth]{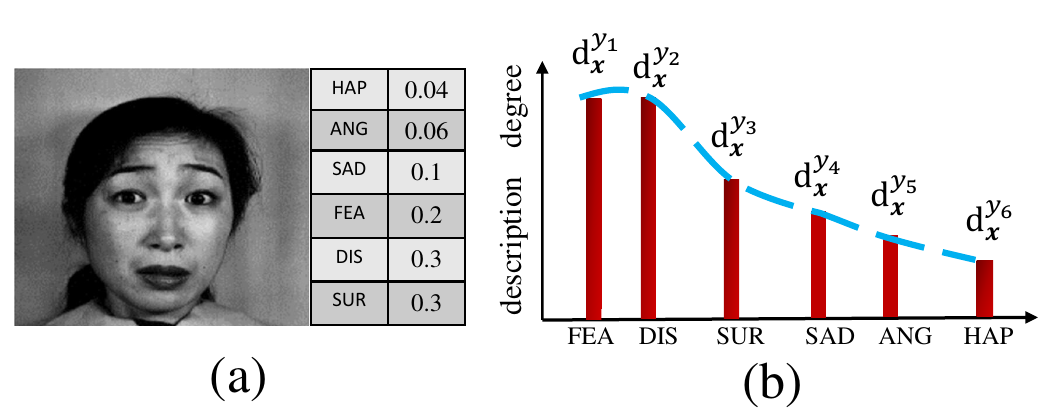}
	\caption{  An illustration of an example from a facial SJAFFE dataset \cite{shih2008performance} annotated with a label distribution. }
	\label{fig1}
\end{figure}

Label Distribution Learning (LDL) \cite{geng2016label} is an innovative machine learning paradigm to tackle the issue of label ambiguity.  Unlike multi-label learning (MLL) \cite{MLL-ZHANG2014}, LDL assigns not only a specific number of labels to each instance but also the importance degree of each label.  For instance, as shown in Fig. \ref{fig1}, an example from a facial emotion dataset\cite{shih2008performance} is displayed, as well as the annotated label distribution. The importance degree is referred to as the label description degree \cite{geng2016label,JiaSLLZ23}, which offers more comprehensive semantic information.  Recent studies have witnessed the significant advancements made by LDL in various practical applications, such as expression recognition \cite{ChenWCSGR20},  facial age estimation \cite{geng2013facial}, image  object detection \cite{xu2023gaussian}, joint acne image grading \cite{WuWLLSCY19}  and head-pose estimation \cite{LiuCBLL19}.

The success of deep learning heavily rely on large-scale and accurately labeled datasets, which are necessary to train very deep neural networks (DNNs) with superior generalization. However, acquiring such labeled data can be an arduous and costly process. Especially, it is more costly to obtain large dataset annotated with label distribution. For instance, considering the RAF-LDL dataset \cite{li2019blended}, 315 trained annotators were employed, and each image is annotated for enough independent times to get the appropriate label distribution. As a result, the conflict emerges prominently when LDL embraces DNNs. A possible  way to address the challenge is to leverage the highly available unlabeled data. In this paper, we present the semi-supervised LDL (SSLDL), which aims to develop an LDL model with a few labeled data and a larger pool of unlabeled data.

Notice that semi-supervised learning (SSL) has already make  significant advancements \cite{BasakY23,FiniAAAMN023}, especially in the era of deep learning. However, SSLDL has not been explored to the same extent. Traditional SSL approaches are mainly designed for SLL or MLL, which often rely on confidence-based pseudo-labeling \cite{jiang2022maxmatch}, \cite{sohn2020fixmatch} and fall in SSLDL because its goal is to predict the whole label distribution, not just the most likely label. Moreover, exiting SSL methods typically ignore the correlation between labels \cite{xu2017incomplete}, potentially hindering their performance for LDL.

To solve the challenging SSLDL problem, we put forward in the paper a novel SSLDL method called RankMatch. It uses an averaging strategy from the ensemble learning \cite{zhou2021ensemble}, taking the mean of predictions from variously augmented images \cite{sohn2020fixmatch} to form a pseudo label distribution. RankMatch approach aims to stabilize the predictions and improve model robustness. Moreover, RankMatch incorporates a pairwise relevance ranking loss to acknowledge and utilize the relationships between labels, aligning the predicted label distributions with the ground truth.  In the theoretical analysis, we establish a generalization bound for RankMatch. Finally, in the experiments, we demonstrate that RankMatch can effectively address the SSLDL problem and outperform existing methods.

In summary, our contributions can be summarized as
\begin{itemize}
	\item To the best of our knowledge, this is the first work employing deep learning to address the SSLDL problem.

	\item We propose the RankMatch approach that leverages inter-label correlations to tackle the SSLDL problem.
	
	\item We establish a theoretical generalization bound for RankMatch and validate its efficacy through extensive experimentation. 
\end{itemize}

\begin{figure*}[!h]
	\centering
	\includegraphics[width=0.51\textwidth]{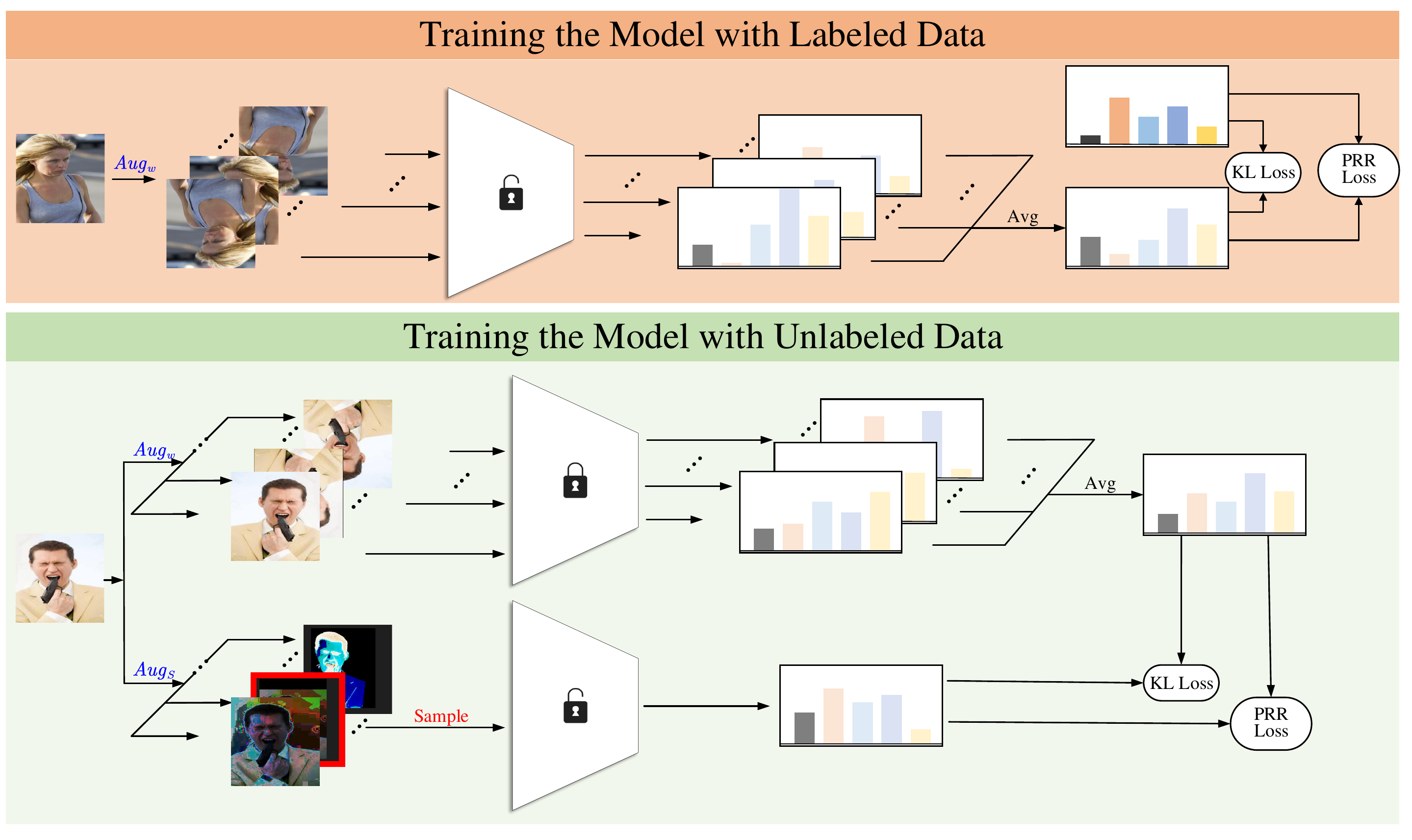}
	\caption{ Overview of the RankMatch Algorithm for Semi-Supervised Label Distribution Learning: The diagram showcases the model training phases, where the locked icon indicates phases with fixed parameters and the unlocked icon represents parameter updates. The top (orange) delineates the process with labeled data, while the bottom (green) details the approach for integrating unlabeled data.}
	\label{overall}
\end{figure*}

\section{RELATED WORK}
\subsection{Label Distribution Learning}
Label distribution learning (LDL) \cite{geng2016label} is an innovative learning approach that assigns a label distribution to each instance and directly learns a mapping from instances to these label distributions. The initial proposal of LDL was aimed at addressing the problem of facial age estimation \cite{geng2013facial}. It  involves the label distributions generated for all age groups, which is considered more advantageous than relying on a single age label. Next, Geng\cite{geng2016label} discovered that in some practical applications, it is more preferable to have distributions spanning all labels rather than associating a single label with an instance\cite{xu2017incomplete}. Such as, in the task of facial emotion recognition, the human emotions are often a blend of multiple emotional states rather than a single one. Because LDL can model the uncertainty of the label space, it has been receiving increasing attention. To name some examples, NASA employs LDL for predicting mineral compositions in Martian meteorite craters \cite{Morrison2018PredictingMM}. In this research, they fine-tuned the LDL algorithm to predict the chemical elements (labels) and their abundances (degrees) for each Martian mineral sample based on crystallographic parameters. The research by \cite{zhou2020facial} addresses the challenge of facial expression variations in depression recognition by framing it as a label distribution learning (LDL) task. Their innovative Deep Joint Label Distribution and Metric Learning (DJ-LDML) approach effectively navigates the nuanced differences in facial expressions, both within the same depression level and across different levels, enhancing the accuracy of depression assessment. In indoor videos, people often remain relatively still for extended periods. Based on this observation, Ling \cite{ling2019indoor} employs label distributions to cover a certain number of crowd count labels, representing the level of description each label provides for video frames, in order to address the crowd counting problem in indoor videos.

\subsection{Semi-supervised Label Distribution Learning}

Lack of sufficient training data with exact labels is still a challenge for label distribution learning. To deal with such problem, many algorithms have been proposed, known as Semi-Supervised Label Distribution Learning (SSLDL). To named some, Hou \cite{hou2017semi}  uses the average of labels from the neighbors of unlabeled data as its label distribution, and then trains the LDL model with both labeled and unlabeled data. Jia \cite{jia2021semi} recovers unknown label distributions by utilizing sample-related information among graph nodes. Liu \cite{liu2022semi} proposed a semi-supervised label distribution learning algorithm based on co-regularization, which employs two different model structures to handle labeled and unlabeled data, demonstrating strong robustness and consistency. 

Although a range of SSLDL approaches have been proposed, they are not end-to-end. Traditional methods typically necessitate manual feature engineering and struggle with large-scale, high-dimensional data, while also inadequately leveraging unlabeled data. In contrast, deep learning is renowned for its ability to autonomously learn complex feature representations and has proven effective in various data-intensive tasks. Thus, we are motivated to explore the potential of deep learning in addressing SSL challenges, with the aim of overcoming the limitations inherent in conventional techniques.

\section{The Method}
\subsection{Problem Statement and Notation}
In semi-supervised label distribution learning (SSLDL), we have a training set denoted by \( \mathcal{D} \), which combines labeled and unlabeled datasets: \( \mathcal{D}_L = \{ (\mathbf{x}_i, \mathbf{d}_i) | i \leq n \} \) for labeled samples and \( \mathcal{D}_U = \{ \mathbf{x}_j | j \leq m \} \) for unlabeled samples. The instance variable is denoted by \( \mathbf{x} \), the particular i-th instance is denoted by \( \mathbf{x}_i \),   \(\mathbf{ d_i} = \{ d^{y_1}_\mathbf{x_i}, d^{y_2}_\mathbf{x_i}, ..., d^{y_c}_\mathbf{x_i} \} \) represents the label distribution for instance \( \mathbf{x}_i \), where c is the  number of label, and
\( d_{\mathbf{x}_i}^{y_c} \) reflects the relevance of label \( y_c \) to the instance, and $\sum_{j=1}^c d_{\mathbf{x}_i}^{y_j}=1$. Our goal is to train a deep neural network (DNN), represented as \( f(\mathbf{x}; \theta) \), to predict these label distributions. The model's output for each label \( y_j \) is normalized using the Softmax function \cite{jang2016categorical} to ensure a proper probability distribution:
\begin{equation}
	h(y_j | {\mathbf{x}}_i ; \theta) = \frac{\exp(f_j(\mathbf{x}_i; \theta))}{\sum_q \exp(f_q(\mathbf{x}_i; \theta))},
\end{equation}
where \( f_j(\mathbf{x}_i; \theta) \) is the raw output of the DNN for label \( y_j \) and the instance \( \mathbf{x}_i \), and the denominator is the sum over all possible labels, ensuring the output for each instance sums to 1 \cite{gao2017deep}. This way, \( h(y_j | x_i; \theta) \) provides a normalized prediction of label \( y_j \)'s relevance for each instance \( \mathbf{x}_i \).
\subsection{The Supervised Loss}
In label distribution learning, we diverge from the traditional binary cross-entropy loss used in multi-label learning \cite{hershey2007approximating} , as LDL requires predicting a range of label intensities rather than separate binary outcomes. We use Kullback-Leibler (KL) divergence \cite{hershey2007approximating}   as the loss function to measure the difference between the predicted and actual label distributions. The supervised loss is  defined  as
\begin{small}
	\begin{equation}
		\mathcal{L}_{s} = \frac{1}{n} \sum_{i=1}^n \sum_{j=1}^q d_{\mathbf{x}_i}^{y_j} \ln \left( \frac{d_{\mathbf{x}_i}^{y_j}}{h(y_j \mid \operatorname{Aug}_w(\mathbf{x}_i))} \right),
	\end{equation}
\end{small}
where \( \operatorname{Aug}_w(\mathbf{x}_i) \) indicates the weak augmentation applied to the \( i \)-th sample, and \( h(y_j \mid \operatorname{Aug}_w(\mathbf{x}_i)) \) is the DNN's predicted probability for label \( y_j \). This loss encourages the model to align closely with the true label distribution.
\subsection{The Unsupervised Consistency Loss}
In semi-supervised label distribution learning, the challenge lies in leveraging both the labeled and the significant volume of unlabeled data effectively. Consistency regularization  emerges as a potent strategy, drawing inspiration from recent advancements in SSL \cite{jiang2022maxmatch} \cite{sohn2020fixmatch} \cite{yang2022survey}  \cite{zhang2021flexmatch}. The principle driving this approach is ensuring that the classifier's output remains stable across different augmentations of the same unlabeled instance, thereby reinforcing the reliability of the label distribution predictions.

To enhance the stability of predictions and tap into the full potential of unlabeled data, we employ an ensemble learning-inspired technique \cite{zhou2021ensemble}. Rather than relying on high-confidence predictions, we average the outputs from multiple weakly augmented \cite{sohn2020fixmatch} variants of the same unlabeled image. This process forms what we call the pseudo-label distribution (PLD) for each unlabeled instance, described as \( \mathbf{p}_{i} \), which aggregates the predictions across augmentations and smooths out anomalies due to random variance in the data augmentation process.
The unsupervised consistency loss, \( \mathcal{L}_{uc} \), is then calculated by contrasting the PLD with the model's predictions for strongly augmented \cite{sohn2020fixmatch} versions of the same instances.  The formula is given by
\begin{small}
	\begin{equation}
		\mathcal{L}_{uc} = \frac{1}{m} \sum_{u=1}^m \sum_{j=1}^q \left( p_{\mathbf{x}_u}^{y_j} \ln \left( \frac{p_{\mathbf{x}_u}^{y_j}}{h\left(y_j \mid \operatorname{Aug}_s(\mathbf{x}_u)\right)} \right) \right),
	\end{equation}
\end{small}
where \( h\left(y_j \mid \operatorname{Aug}_s(\mathbf{x}_u)\right) \) represents the prediction for label \( y_j \) post strong augmentation \cite{sohn2020fixmatch}. This loss function  plays a pivotal role in SSLDL by guiding the model to learn from the structure within the data, even when explicit labels are not available.
\subsection{ The Pairwise Relevance Ranking Loss}
The supervised loss and the unsupervised consistency loss both treat the predicted results and ground-truth (or PLD) as multiple independent prediction tasks, thereby overlooking the inter-label correlation \cite{xu2017incomplete}, which may lead to a decrease in performance.  In LDL, a sample is assigned multiple label description degree , and these description degree are often not completely independent of each other \cite{jiaLDLLC}. The correlation between the description degrees can be either positive or negative. For example, if an image $\mathbf{x}$ has a label distribution of $d_{\mathbf{x}}^{y_1}=0.4$ and $d_{\mathbf{x}}^{y_2}=0.2$, we consider labels $y_1$ and $y_2$ to be negatively correlated. Similarly, if the labels have a distribution of $d_{\mathbf{x}}^{y_1}=0.4$ and $d_{\mathbf{x}}^{y_2}=0.4$, we consider labels $y_1$ and $y_2$ to be positively correlated.  This pairwise ranking relationship implicitly expresses the label correlation between label distributions.  

To tackle this challenge, we introduce a pairwise relevance ranking (PRR) loss $\mathcal{L}_{PRR} $ to align this inherent semantic structure.  For labeled data, we aim for a strict alignment between the ranking of predicted label distributions and the ground-truth. This means that we not only need to align the ranking relationships between label descriptions but also maintain the margin with the ground-truth. Additionally, for certain "close" description degrees, studying their ranking is not meaningful.  For instance, consider a scenario where the label description degrees \( d_{\mathbf{x}}^{y_i} \) and \( d_{\mathbf{x}}^{y_k} \) are 0.32 and 0.33, respectively. The negligible discrepancy between these two values could be attributed to variations in annotation. Consequently, we opt not to adjust their ranking order to account for such minor differences, which may not reflect actual dissimilarities in label importance. Simplifying our notation, let \( h_j(\mathbf{x}_i) \) represent the predicted degree of relevance for the j-th label after applying a weak augmentation \( \operatorname{Aug}_w \) to the i-th instance. The \( \mathcal{L}_{PRR_L} \) loss is then defined as follows:
\begin{small}
	\begin{equation}
		\begin{aligned}
			\mathcal{L}_{PRR_L} = &\sum_{1<j<k<q} 
			I(d_{\mathbf{x}_i}^{y_j}, d_{\mathbf{x}_i}^{y_k}) \cdot \max(0, \delta - (h_j(\mathbf{x}_i) - h_k(\mathbf{x}_i))) \\
			&+ I(d_{\mathbf{x}_i}^{y_k}, d_{\mathbf{x}_i}^{y_j}) \cdot \max(0, \delta - (h_k(\mathbf{x}_i) - h_j(\mathbf{x}_i))),
		\end{aligned}
	\end{equation}
\end{small}
\begin{figure}[!h]
	\centering
	\includegraphics[width=0.5\textwidth]{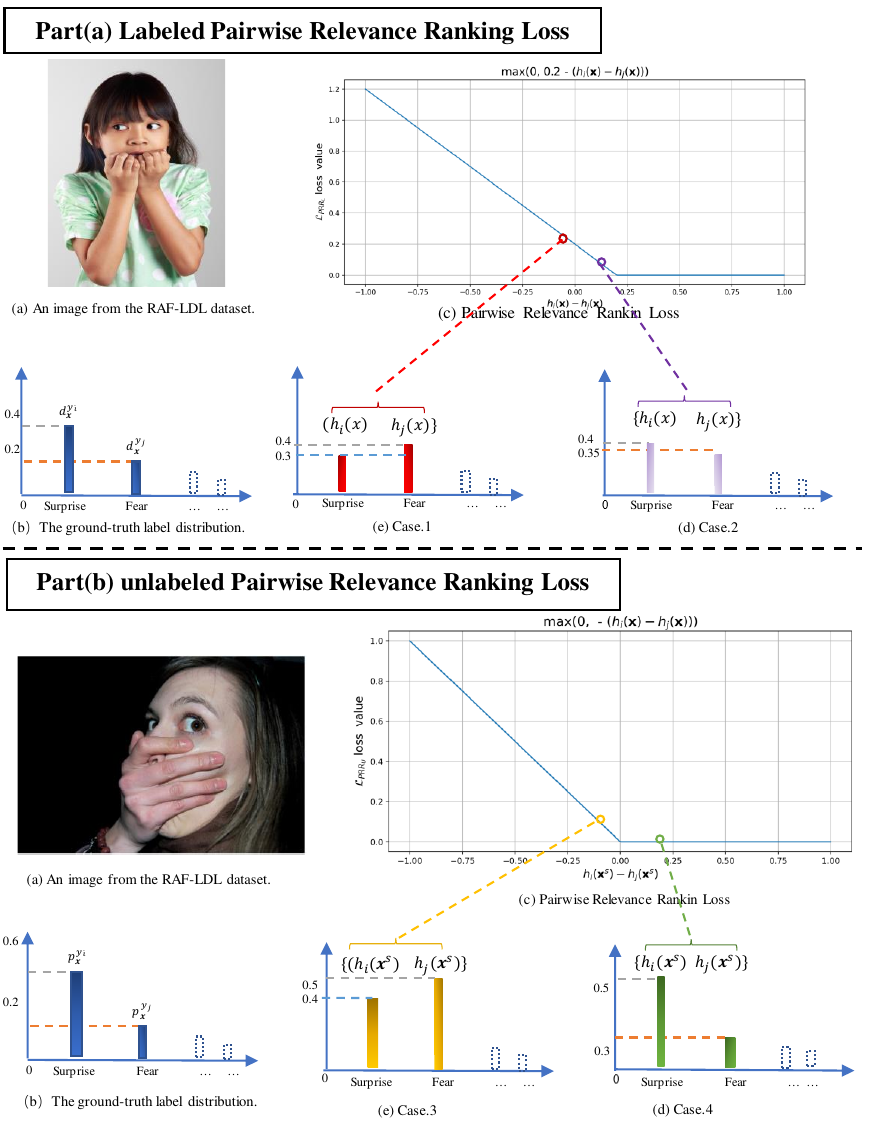}
	\caption{An examples to illustrate  the $\mathcal{L}_{PRR} $ loss.}
	\label{rankingloss}
\end{figure}
Fig. \ref{rankingloss}, Part (a), presents an image from the RAF-LDL dataset and its label distribution, illustrating the application of the \(\mathcal{L}_{PRR_L}\) loss. Here, \( \delta = d_{\mathbf{x}_i}^{y_k} - d_{\mathbf{x}_i}^{y_j} \) and the function $I(d_{\mathbf{x}_i}^{y_j},d_{\mathbf{x}_i}^{y_k})$ is an indicator that outputs 1 if the first label's degree is greater than the second's and their difference is significant, i.e., \( d_{\mathbf{x}_i}^{y_j} > d_{\mathbf{x}_i}^{y_k} \) and \( |d_{\mathbf{x}_i}^{y_j} - d_{\mathbf{x}_i}^{y_k}| > t \). The loss comes into play in two key scenarios: Case 1, when the model's predicted ranking of labels is incorrect, and Case 2, when the ranking is correct but the margin does not align with the ground truth. Both cases indicate opportunities for the model to learn and adjust its predictions.

In the unsupervised component of our model, we confront the absence of ground-truth labels by employing pseudo-label distributions (PLDs) as a stand-in during training. Recognizing that PLDs may not always be precise, we focus on aligning the predicted pairwise relevance rankings of label descriptions to mitigate the potential for overfitting and to correct inaccuracies inherent in SSL.  We define the unsupervised pairwise relevance ranking loss, \( \mathcal{L}_{PRR_u} \), where \( h_j(\mathbf{x}_i^s) \) denotes the predicted relevance of the j-th label after strong augmentation, \( \operatorname{Aug}_s \), is applied to the i-th instance. The loss function is as follows:
\begin{small}
	\begin{equation}
		\begin{aligned}
			\mathcal{L}_{PRR_u} = &\sum_{1<j<k<q} I(p_{\mathbf{x}_i}^{y_j}, p_{\mathbf{x}_i}^{y_k}) \cdot \max(0, -(h_j(\mathbf{x}_i^s) - h_k(\mathbf{x}_i^s)))\\
			& + I(p_{\mathbf{x}_i}^{y_k}, p_{\mathbf{x}_i}^{y_j}) \cdot \max(0, -(h_k(\mathbf{x}_i^s) - h_j(\mathbf{x}_i^s))),
		\end{aligned}
	\end{equation}
\end{small}
where the indicator function, $I(p_{\mathbf{x}_i}^{y_j},p_{\mathbf{x}_i}^{y_k})$, outputs 1 if the pseudo-label of one label is greater than the other and their difference is substantial, specifically when \( p_{\mathbf{x}_i}^{y_j} > p_{\mathbf{x}_i}^{y_k} \) and the difference \( |p_{\mathbf{x}_i}^{y_j} - p_{\mathbf{x}_i}^{y_k}| \) exceeds a threshold \( t \); otherwise, it outputs 0.  This loss addresses the scenario where the model's ranking of label predictions is inaccurate, as illustrated in Fig. \ref{rankingloss}, Part (b). Here, we see an image from the RAF-LDL dataset and its associated pseudo-label distribution. For example, when the PLD for surprise (\( p_{\mathbf{x}}^{y_i} \)) is 0.6 and for fear (\( p_{\mathbf{x}}^{y_j} \)) is 0.2, the \( \mathcal{L}_{PRR} \) loss is activated as \( \max(0, -(h_i(\mathbf{x}) - h_j(\mathbf{x}))) \), emphasizing the need for the model to correct the predicted rankings to reflect the pseudo-labels more accurately.

In conclusion, the RankMatch algorithm, as illustrated in Fig. \ref{overall}, harnesses a dual-phase training strategy for semi-supervised label distribution learning.  It differentiates between the treatment of labeled and unlabeled data, refining the model with fixed parameters for the former and updating parameters for the latter. Our tailored loss function combines supervised and unsupervised ranking losses under the PRR framework, with a lambda (\( \lambda \)) coefficient to balance their influence. Thus, the total loss is defined as \( \text{loss} = \mathcal{L}_s + \mathcal{L}_{uc} + \lambda (\mathcal{L}_{PRR_L} + \mathcal{L}_{PRR_u}) \), streamlining the model’s learning from both labeled and unlabeled datasets.
\subsection{Theoretical Analysis}
In this section, we establish a theoretical foundation for our RankMatch algorithm within the realm of Semi-Supervised Label Distribution Learning (SSLDL) by defining a generalization bound.
\subsubsection{Generalization Bound}
\begin{theorem}
	For any function \( f \) in \( \mathcal{F}_{\beta, \nu} \) \cite{JiangLCHXYCH23}, the following inequality is satisfied with probability at least \( 1-\delta \) over the random selection of \( \mathcal{D}_L \) and \( \mathcal{D}_U \):
	\begin{small}
		\begin{equation*}
			\begin{aligned}
				R_{ \text {-KL }}(f) & \leq  C_1 \cdot \widehat{R}_{\mathcal{D}_U}(f)+C_2\left(1+\frac{C_0}{2}\right) \cdot \widehat{R}_{\mathcal{D}_L}(f) \\
				& +C_1(\frac{16 K q}{\sqrt{m}} \cdot \frac{1-\varepsilon}{\varepsilon}\left[1+3 \sqrt{W_g \cdot \log \left(C_M m\right)}\right]\\
				& +3 \sqrt{\frac{\log \frac{4}{\delta}}{2 m}})+\frac{3 C_0 C_2}{2} \cdot \Psi_{n, q, \delta}\left(\mathcal{F}_{\beta, m}\right),
				\label{Theorem1}
			\end{aligned}
		\end{equation*}	
	\end{small}
	where  \( \hat{R}_{D_L}(f) \) and the latter by \( \hat{R}_{D_U}(f) \) denote   the empirical risk on labeled and unlabeled datasets,  \( C_0 \) is a universal constant, \( C_1 \) and \( C_2 \) are constants determined by the neural network capacity and the structure of the data distribution, $\varepsilon \in (0, 0.5)$. And the last term is the complexity terms and they are dominated by the upper bound of the Rademacher complexity. Specifically,  \( \Psi_{n, q, \delta}(\mathcal{F}_{\beta, m}) \) are given by
	\begin{small}
		\begin{equation*}
			\begin{aligned}
				\Psi_{n, q, \delta}(\mathcal{F}_{\beta, \nu}) = &2 (\sqrt{q} \cdot \log^{3/2}(n q e) \cdot \psi_{n, q}(\mathcal{F}_{\beta, \nu})\\
				& + \frac{1}{\sqrt{n}} ) + \frac{\log(q e)}{n} \left(\log \frac{2}{\delta} + \log(\log n) \right).
			\end{aligned}
		\end{equation*}
		\begin{equation*}
			\psi_{n,q}(\mathcal{F}_{\beta, m}) = \frac{4}{\sqrt{n \cdot q}} + \frac{12}{\sqrt{n \cdot q}} \sqrt{W \cdot \log(C_N \sqrt{n q})},
		\end{equation*}
	\end{small}
\end{theorem}
The left-hand side of the inequality, \( R_{KL}(f) \), represents the expected robust risk when measured with the KL divergence. The right-hand side aggregates the empirical risks and a series of complexity terms, encapsulating the Rademacher complexity.  The detailed proof of Theorem 1, which includes the definitions and derivations of the Rademacher complexity terms as well as the constants \( C_0 \), \( C_1 \), \( C_2 \),\( C_M \), \( C_N \) and the set \( \mathcal{B}_{sem}(x_i^u) \), is located in the appendix. This proof lays the groundwork for asserting the model's efficacy and robustness in learning from both labeled and unlabeled data in an SSLDL setting.

\begin{Remark}
	Theorem \ref{Theorem1}   implies that the third term is of
	order $O\left(K \sqrt{\frac{W_g \log m}{m}}\right)$, and the fourth term is of order $\Psi_{n, q, \delta}\left(\mathcal{F}_{\beta, \nu}\right)$,  more specifically, in $O\left(\log ^2\left(n q\right) \sqrt{\frac{W}{n}}\right)$. Therefore, the order of the generalization gap is $O\left(K \sqrt{\frac{W_g \cdot \log m}{m}}+\log ^2\left(n q\right) \sqrt{\frac{W}{n}}\right)$.
\end{Remark}
\subsection{Experiments}

\subsubsection{ Experimental Configurations}
\textbf{Experimental Datasets }:In this paper, we validate our approach using four distinct real-world datasets. The details of these datasets are as follows:

Twitter-LDL \cite{yang2017learning}: A large-scale Visual Sentiment Distribution dataset was constructed from Twitter, encompassing eight distinct emotions {Amusement, Anger, Awe, Contentment, Disgust, Excitement, Fear, Sadness}. Approximately 30,000 images were collected by searching various emotional keywords, such as "sadness," "heartbreak," and "grief." Subsequently, eight annotators were hired to label this dataset. The resulting Twitter LDL dataset comprises 10,045 images.

Flickr-LDL \cite{yang2017learning}: A subset of the Flickr dataset \cite{borth2013large}, unlike other datasets that searched for images using emotional terms, the Flickr dataset collected 1,200 pairs of adjective-noun pairs, resulting in 500,000 images. We employed 11 annotators to label this subset with tags for eight common emotions. In the end, the Flickr LDL was created, containing 10,700 images, with roughly equal quantities for each class.

Emotion6 \cite{peng2015mixed}: Emotion6: We collected 1,980 images from Flickr \cite{borth2013large} using six category keywords and synonyms as search terms for Emotion6. A total of 330 images were collected for each category, and each image was assigned to only one category (dominant emotion). Emotion6 represents the emotions related to each image in the form of a probability distribution, consisting of 7 bins, including Ekman's 6 basic emotions \cite{carroll1996facial} and neutral. 

RAF-LDL \cite{li2019blended}: RAF-LDL is a multi-label distribution facial expression dataset, comprising approximately 5,000 diverse facial images downloaded from the internet. These images exhibit variations in emotion, subject identity, head pose, lighting conditions, and occlusions.  During annotation, 315 well-trained annotators are employed to ensure each image can be annotated enough independent times. And images with multi-peak label distribution are selected out to constitute the RAF-LDL.

\textbf{Comparing methods}  In order to assess the effectiveness of the proposed approach, we benchmark it against four sets of methods: 1) The first group consists of two deep learning SSLDL algorithms that we introduced, named FixMatch-LDL \cite{sohn2020fixmatch}  and MixMatch-LDL \cite{peng2015mixed}. Since there are currently no open-source semi-supervised LDL works in deep learning, these two algorithms were developed by us, based on the current most effective two deep learning SSL algorithms.
2) The second group of algorithms is a deep learning SSLDL algorithm based on the dual-network concept \cite{ChenYZ021}, which we named GCT-LDL. The core idea involves mutual supervision of the outputs from two independent networks using unlabeled data.  3) The third group consists of traditional SSLDL algorithms, referred to as SA-LDL \cite{hou2017semi}. Since SA-LDL is an SSLDL algorithm designed for tabular data, we needed to perform feature engineering on image data, the details can be found in appendix.  4) The fourth category consists of existing LDL algorithms. As there is currently only one open-source SSLDL algorithm, which is SA-LDL \cite{hou2017semi}, we compared it with some state-of-the-art LDL algorithms. In this regard, we selected four state-of-the-art LDL algorithms: Adam-LDL-SCL \cite{jia2019label}, sLDLF \cite{shen2017label}, DF-LDL \cite{gonzalez2021decomposition}, and LDL-LRR \cite{jia2021label}. These methods are designed for fully labeled LDL algorithms and may not be suitable for the SSLDL problem. All algorithm details and configurations can be found in the appendix.

\textbf{Implementation} Following \cite{cole2021multi}, we employ ResNet-50 \cite{he2016deep} pre-trained on ImageNet \cite{krizhevsky2012imagenet} for training the classification model.  For training images, we adopt standard flip-and-shift strategy \cite{sohn2020fixmatch} for weak data augmentation, and RandAugment \cite{cubuk2020randaugment}  and Cutout \cite{devries2017improved} for strong data augmentation. We employ AdamW \cite{you2019large} optimizer and one-cycle policy scheduler \cite{hannan2021state} to train the model with maximal learning rate of 0.0001. For all datasets, the number of epochs is set as 30 and the batch size is set as 32. Furthermore, we perform exponential moving average (EMA) \cite{klinker2011exponential} for the model parameter $\theta$ with a decay of 0.98.  We adjust the parameter \( \lambda \) across a range of values, specifically \( \{0.005, 0.01, 0.05, 0.1\} \). We perform all experiments on GeForce RTX 3090 GPUs. The random seed is set to 1 for all experiments.

\textbf{Evaluation Metrics}: In evaluating LDL methods, we employ six distinct metrics \cite{geng2016label}: Chebyshev, Clark, and Canberra distances, along with Kullback-Leibler divergence, where lower values are preferable, and Intersection and Cosine similarities, where higher values indicate better performance. Details of the evaluation metrics are provided in the Appendix.

\subsubsection{Comparative Experiment Analysis}

We employed a range of labeled data proportions (10\%, 20\%, and 40\%) to simulate varying levels of label availability, a critical factor in semi-supervised learning scenarios. Our evaluation metrics included Canberra and Clark distances, Intersection and Cosine similarities, and KL divergence—providing a multifaceted evaluation of predictive performance.

The results, as displayed in Table. \ref{zhushiyan1} and Table. \ref{zhushiyan2}, reveal a consistent pattern of superior performance by the RankMatch algorithm. Particularly, it exhibits lower scores in Canberra and Clark distances across all datasets and sample proportions, signifying its ability to minimize prediction errors effectively. This is indicative of RankMatch's capacity to capture the intricate label relationships even with limited labeled information. In terms of Intersection and Cosine similarities, RankMatch consistently outperformed other methods, particularly at lower proportions of labeled data. This trend underscores its efficient utilization of the available labeled data and its potent semi-supervised learning strategy, which exploits the unlabeled data effectively. Furthermore, the leading results for each dataset and sample proportion are highlighted in bold, illustrating instances where RankMatch not only performed better compared to other methods but also when it outshined its own performance across different datasets. These leading performances substantiate the model's versatility and its suitability for diverse label distribution learning tasks.

In conclusion, the comparative experiment analysis solidifies RankMatch as a robust and versatile algorithm for semi-supervised label distribution learning, capable of delivering outstanding performance across various datasets and under different label availability scenarios.

\begin{table*}[]\centering\scriptsize
	\setlength{\tabcolsep}{5pt}
	\renewcommand{\arraystretch}{1}
	\begin{tabular}{@{} lllllll|lllllll @{}}
		\toprule
		& \multicolumn{3}{c}{emotion6  Canberra $\downarrow$} &\multicolumn{3}{c|}{flickr Canberra   $\downarrow$} &  & \multicolumn{3}{c}{emotion6 Clark   $\downarrow$} & \multicolumn{3}{c}{flickr Clark   $\downarrow$} \\ \hline
		Method & 10\% & 20\% & 40\% & 10\% & 20\% & \multicolumn{1}{c|}{40\%} & Method & 10\% & 20\% & 40\% & 10\% & 20\% & 40\% \\ 
		\midrule
		Rankmatch & \textbf{3.3902} & \textbf{3.3176} & 3.2504 & \textbf{4.4060} & \textbf{3.9964} & \multicolumn{1}{c|}{\textbf{3.9013}} & Rankmatch & \textbf{1.5298} & \textbf{1.5050} & \textbf{1.4834} & \textbf{1.8189} & \textbf{1.7051} & \textbf{1.6737} \\
		fixmatch-LDL & 3.5080 & 3.5680 & 3.6050 & 5.5570 & 5.5310 & \multicolumn{1}{c|}{5.4350} & fixmatch-LDL & 1.5950 & 1.6230 & 1.6390 & 2.2220 & 2.2110 & 2.1910 \\
		mixmatch-LDL & 3.6080 & 3.4860 & 3.4880 & 5.6450 & 5.5026 & \multicolumn{1}{c|}{5.5750} & mixmatch-LDL & 1.6240 & 1.5810 & 1.5840 & 2.2330 & 2.1996 & 2.2160 \\
		GCT-LDL & 3.5980 & 3.5490 & 3.6410 & 5.5860 & 5.5872 & \multicolumn{1}{c|}{5.5260} & GCT-LDL & 1.6090 & 1.6050 & 1.6390 & 2.2200 & 2.2238 & 2.2080 \\
		SALDL & 3.4836 & 3.3737 & \textbf{3.1931} & 5.4612 & 4.7789 & \multicolumn{1}{c|}{4.8199} & SALDL & 1.6019 & 1.5751 & 1.5100 & 2.1967 & 2.0369 & 2.0446 \\
		sLDLF & 4.4164 & 4.3398 & 4.1322 & 6.2280 & 6.1238 & \multicolumn{1}{c|}{6.2589} & sLDLF & 1.8922 & 1.8566 & 1.8049 & 2.3722 & 2.3436 & 2.3761 \\
		DF-LDL & 4.2427 & 4.0717 & 3.7221 & 5.5348 & 5.5549 & \multicolumn{1}{c|}{5.5207} & DF-LDL & 1.8217 & 1.7746 & 1.6781 & 2.2253 & 2.2072 & 2.1992 \\
		LDL-LRR & 4.6528 & 4.0496 & 3.7719 & 5.6325 & 5.4988 & \multicolumn{1}{c|}{5.4319} & LDL-LRR & 1.9899 & 1.7745 & 1.6953 & 2.2285 & 2.2026 & 2.1919 \\
		Adam-LDL-SCL & 4.7403 & 4.4267 & 4.1714 & 6.2821 & 6.2436 & \multicolumn{1}{c|}{5.8719} & Adam-LDL-SCL & 2.0334 & 1.8864 & 1.8073 & 2.3857 & 2.3666 & 2.2811 \\ 
		\toprule
		& \multicolumn{3}{c}{twitter Canberra $\downarrow$} & \multicolumn{3}{c|}{RAF Canberra $\downarrow$} &  & \multicolumn{3}{c}{twitter Clark $\downarrow$} & \multicolumn{3}{c}{RAF Clark $\downarrow$} \\ \hline
		Method & 10\% & 20\% & 40\% & 10\% & 20\% & \multicolumn{1}{c|}{40\%} & Method & 10\% & 20\% & 40\% & 10\% & 20\% & 40\% \\ 
		\midrule
		Rankmatch & \textbf{3.7370} & \textbf{3.6962} & \textbf{3.2913} & \textbf{3.0178} & \textbf{2.9358} & \multicolumn{1}{c|}{\textbf{2.8341}} & Rankmatch & \textbf{1.6480} & \textbf{1.6190} & \textbf{1.5138} & \textbf{1.4506} & \textbf{1.4190} & \textbf{1.3843} \\
		fixmatch-LDL & 6.1750 & 6.0060 & 5.8340 & 3.1220 & 3.0920 & \multicolumn{1}{c|}{3.0770} & fixmatch-LDL & 2.3830 & 2.3310 & 2.2820 & 1.5130 & 1.5060 & 1.5050 \\
		mixmatch-LDL & 6.3530 & 6.2489 & 6.2960 & 3.1580 & 3.1111 & \multicolumn{1}{c|}{3.0630} & mixmatch-LDL & 2.4280 & 2.4034 & 2.4150 & 1.5150 & 1.5020 & 1.4870 \\
		GCT-LDL & 6.3010 & 6.3078 & 6.2380 & 3.1920 & 3.1260 & \multicolumn{1}{c|}{3.1470} & GCT-LDL & 2.4170 & 2.4216 & 2.4060 & 1.5350 & 1.5170 & 1.5290 \\
		SALDL & 5.0380 & 4.0868 & 4.0742 & 3.1947 & 3.1415 & \multicolumn{1}{c|}{3.0527} & SALDL & 2.1288 & 1.8938 & 1.8964 & 1.5445 & 1.5288 & 1.5035 \\
		sLDLF & 5.3084 & 6.0008 & 6.1910 & 4.0586 & 4.1705 & \multicolumn{1}{c|}{4.1189} & sLDLF & 2.1480 & 2.3384 & 2.3746 & 1.9300 & 1.9645 & 1.9750 \\
		DF-LDL & 6.4184 & 6.3120 & 6.2588 & 3.3281 & 3.3865 & \multicolumn{1}{c|}{3.3582} & DF-LDL & 2.4313 & 2.4108 & 2.4033 & 1.6071 & 1.6229 & 1.6138 \\
		LDL-LRR & 6.4215 & 6.3295 & 6.2905 & 3.8677 & 4.0116 & \multicolumn{1}{c|}{4.1890} & LDL-LRR & 2.4429 & 2.4223 & 2.4121 & 1.7907 & 1.8298 & 1.8919 \\
		Adam-LDL-SCL & 6.7296 & 6.2436 & 6.4749 & 3.792 & 3.9255 & \multicolumn{1}{c|}{4.132} & Adam-LDL-SCL & 2.5184 & 2.5165 & 2.4501 & 1.75 & 1.7983 & 1.8634 \\ \hline
		
	\end{tabular}
	\caption{ Performance metrics of RankMatch and benchmark semi-supervised label distribution learning algorithms on Emotion6, Flickr, RAF, and Twitter datasets. Results are evaluated at different training sample proportions: 10\%, 20\%, and 40\%. Metrics are shown for Canberra and Clark distances, with lower scores denoting superior model performance. }
	\label{zhushiyan1}
\end{table*}

\begin{table*}[]\centering\scriptsize
	\setlength{\tabcolsep}{5pt}
	\renewcommand{\arraystretch}{1}
	\begin{tabular}{@{} lllllll|lllllll @{}}
		\toprule
		& \multicolumn{3}{c}{emotion6 Intersection   $\uparrow$} & \multicolumn{3}{c|}{flickr Intersection   $\uparrow$} &  & \multicolumn{3}{c}{emotion6  cosine $\uparrow$} & \multicolumn{3}{c}{flickr cosine   $\uparrow$} \\ \hline
		Method & 10\% & 20\% & 40\% & 10\% & 20\% & 40\% & Method & 10\% & 20\% & 40\% & 10\% & 20\% & 40\% \\ 
		\midrule
		Rankmatch & \textbf{0.6735} & \textbf{0.6832} & \textbf{0.6940} & \textbf{0.6921} & \textbf{0.7073} & \multicolumn{1}{l|}{\textbf{0.7151}} & Rankmatch & \textbf{0.8121} & \textbf{0.8257} & \textbf{0.8331} & \textbf{0.8489} & \textbf{0.8614} & \textbf{0.8679} \\
		fixmatch-LDL & 0.6638 & 0.6797 & 0.6916 & 0.6857 & 0.7042 & \multicolumn{1}{l|}{0.7119} & fixmatch-LDL & 0.8079 & 0.8200 & 0.8312 & 0.8487 & 0.8573 & 0.8673 \\
		mixmatch-LDL & 0.6372 & 0.6418 & 0.6496 & 0.6639 & 0.6686 & \multicolumn{1}{l|}{0.6831} & mixmatch-LDL & 0.7585 & 0.7863 & 0.7901 & 0.7888 & 0.8381 & 0.8468 \\
		GCT-LDL & 0.6116 & 0.6602 & 0.6770 & 0.6639 & 0.6879 & \multicolumn{1}{l|}{0.6863} & GCT-LDL & 0.7530 & 0.8017 & 0.8134 & 0.8313 & 0.8508 & 0.8531 \\
		SALDL & 0.6457 & 0.6612 & 0.6723 & 0.5559 & 0.5108 & \multicolumn{1}{l|}{0.5091} & SALDL & 0.7784 & 0.7874 & 0.7981 & 0.7361 & 0.6643 & 0.6624 \\
		sLDLF & 0.5935 & 0.5861 & 0.6162 & 0.4813 & 0.4750 & \multicolumn{1}{l|}{0.4616} & sLDLF & 0.7037 & 0.6980 & 0.7350 & 0.6276 & 0.6066 & 0.5897 \\
		DF-LDL & 0.5057 & 0.5461 & 0.6353 & 0.4173 & 0.4176 & \multicolumn{1}{l|}{0.4169} & DF-LDL & 0.6035 & 0.6470 & 0.7689 & 0.5436 & 0.5539 & 0.5569 \\
		LDL-LRR & 0.3721 & 0.6213 & 0.6626 & 0.5322 & 0.5519 & \multicolumn{1}{l|}{0.5600} & LDL-LRR & 0.4604 & 0.7362 & 0.7905 & 0.7020 & 0.7316 & 0.7399 \\
		Adam-LDL-SCL & 0.3409 & 0.5627 & 0.604 & 0.4724 & 0.3933 & \multicolumn{1}{l|}{0.4628} & Adam-LDL-SCL & 0.4311 & 0.6670 & 0.7144 & 0.6104 & 0.4888 & 0.6166 \\ \hline
		\toprule
		& \multicolumn{3}{c}{twitter Intersection $\uparrow$} & \multicolumn{3}{c|}{RAF Intersection $\uparrow$} &  & \multicolumn{3}{c}{twitter cosine $\uparrow$} & \multicolumn{3}{c}{RAF cosine $\uparrow$} \\ \hline
		Method & 10\% & 20\% & 40\% & 10\% & 20\% & 40\% & Method & 10\% & 20\% & 40\% & 10\% & 20\% & 40\% \\ 
		\midrule
		Rankmatch & \textbf{0.7036} & \textbf{0.7190} & \textbf{0.7316} & \textbf{0.6551} & \textbf{0.6813} & \multicolumn{1}{l|}{\textbf{0.7044}} & Rankmatch & \textbf{0.8544} & \textbf{0.8698} & \textbf{0.8790} & \textbf{0.7901} & \textbf{0.8140} & \textbf{0.8375} \\
		fixmatch-LDL & 0.7009 & 0.7147 & 0.7283 & 0.6570 & 0.6760 & \multicolumn{1}{l|}{0.6987} & fixmatch-LDL & 0.8517 & 0.8647 & 0.8758 & 0.7881 & 0.8123 & 0.8311 \\
		mixmatch-LDL & 0.6819 & 0.6806 & 0.6986 & 0.6133 & 0.6381 & \multicolumn{1}{l|}{0.6534} & mixmatch-LDL & 0.8463 & 0.8552 & 0.8602 & 0.7536 & 0.7680 & 0.7820 \\
		GCT-LDL & 0.6787 & 0.7018 & 0.7102 & 0.6321 & 0.6669 & \multicolumn{1}{l|}{0.6910} & GCT-LDL & 0.8499 & 0.8587 & 0.8716 & 0.7660 & 0.7977 & 0.8181 \\
		SALDL & 0.6632 & 0.5724 & 0.5687 & 0.6298 & 0.6504 & \multicolumn{1}{l|}{0.6708} & SALDL & 0.8479 & 0.7612 & 0.7615 & 0.7711 & 0.7938 & 0.8135 \\
		sLDLF & 0.6487 & 0.5652 & 0.5336 & 0.2433 & 0.2315 & \multicolumn{1}{l|}{0.2199} & sLDLF & 0.8002 & 0.7454 & 0.6988 & 0.3262 & 0.3506 & 0.3459 \\
		DF-LDL & 0.3541 & 0.3536 & 0.3505 & 0.7022 & 0.7083 & \multicolumn{1}{l|}{0.7085} & DF-LDL & 0.5069 & 0.5233 & 0.5209 & 0.8427 & 0.8492 & 0.8470 \\
		LDL-LRR & 0.5746 & 0.5904 & 0.5979 & 0.5649 & 0.5389 & \multicolumn{1}{l|}{0.4411} & LDL-LRR & 0.7767 & 0.8027 & 0.8125 & 0.7253 & 0.6938 & 0.5757 \\
		Adam-LDL-SCL & 0.5488 & 0.5828 & 0.5200 & 0.6177 & 0.5768 & \multicolumn{1}{l|}{0.4843} & Adam-LDL-SCL & 0.7163 & 0.7661 & 0.7403 & 0.7717 & 0.7337 & 0.6191 \\ \hline
	\end{tabular}
	\caption{ Performance metrics of RankMatch and benchmark semi-supervised label distribution learning algorithms on Emotion6, Flickr, RAF, and Twitter datasets. Results are evaluated at different training sample proportions: 10\%, 20\%, and 40\%. Metrics are shown for Canberra and Clark distances, with lower scores denoting superior model performance.  }
	\label{zhushiyan2}
\end{table*}

\begin{table*} 
	
	\centering\scriptsize
	\setlength{\tabcolsep}{4pt}
	\begin{tabular}{@{}llllllll|l l l l l l l l @{}}
		\toprule
		& & Che. $\downarrow$ & Cla. $\downarrow$ & Can. $\downarrow$ & KL $\downarrow$ & Cos. $\uparrow$ & Int. $\uparrow$& & &  Che. $\downarrow$ & Cla. $\downarrow$ & Can. $\downarrow$ & KL $\downarrow$ & Cos.  $\uparrow$& Int. $\uparrow$\\
		\midrule
		\multirow{3}{*}{Emotion6}
		& pretrain      & 0.2504          & 1.6524          & 3.6893          & 0.4642          & 0.7930           & 0.6557 & \multirow{3}{*}{Twitter}
		& pretrain                               & 0.2538          & 2.4630           & 6.4139          & 0.7908          & 0.8502          & 0.7010 \\
		& + PRR loss     & 0.2464          & 1.6250           & 3.6161          & 0.4456          & 0.8010           & 0.6613 & & + PRR loss     & 0.2549          & 2.1118          & 5.2786          & 0.7157          & 0.8501          & 0.6972 \\
		& + consistency & \textbf{0.2186} & \textbf{1.6028} & \textbf{3.4761} & \textbf{0.3776} & \textbf{0.8349} & \textbf{0.6982} & & + consistency & \textbf{0.2262} & \textbf{1.7382} & \textbf{4.0088} & \textbf{0.6232} & \textbf{0.8799} & \textbf{0.7369} \\
		\midrule
		\multirow{3}{*}{Flickr}
		& pretrain      & 0.2411          & 2.2594          & 5.6885          & 0.5371          & 0.8427          & 0.6873 & \multirow{3}{*}{RAF}
		& pretrain      & 0.2938          & 1.5412          & 3.2060           & 0.5146          & 0.7687          & 0.6411 \\
		& + PRR loss     & 0.2421          & 2.1450           & 5.3781          & 0.5281          & 0.8437          & 0.6858          & & + PRR loss      & 0.2888          & 1.5305          & 3.1878          & 0.5010           & 0.7731          & 0.6428 \\
		& + consistency & \textbf{0.2184} & \textbf{2.0158} & \textbf{4.9008} & \textbf{0.5227} & \textbf{0.8714} & \textbf{0.7208} & & + consistency & \textbf{0.2341} & \textbf{1.4914} & \textbf{3.0459} & \textbf{0.3464} & \textbf{0.8476} & \textbf{0.7194} \\ \hline
	\end{tabular}	\caption{Ablation Results on 4 Datasets. }
	\label{xiaorongshiyan}
\end{table*}

\begin{figure*}[!h]
	\centering
	\begin{subfigure}[b]{0.2\textwidth}
		\includegraphics[scale=0.2, trim={0cm 0 1.5cm 0}, clip]{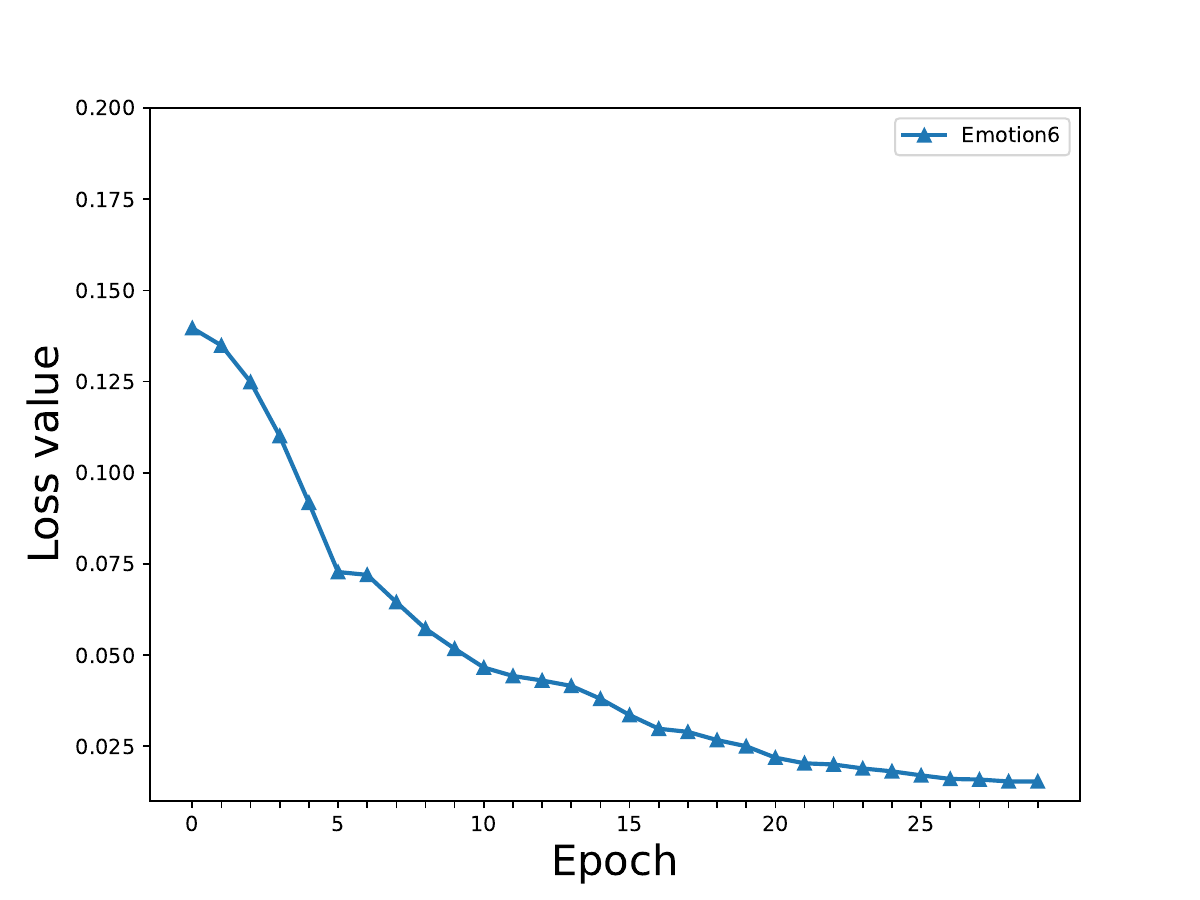}
		\caption{Emotion6}
		\label{fig:emotion6_conver}
	\end{subfigure}
	\hfill
	\begin{subfigure}[b]{0.2\textwidth}
		\includegraphics[scale=0.2, trim={0cm 0 1.5cm 0}, clip]{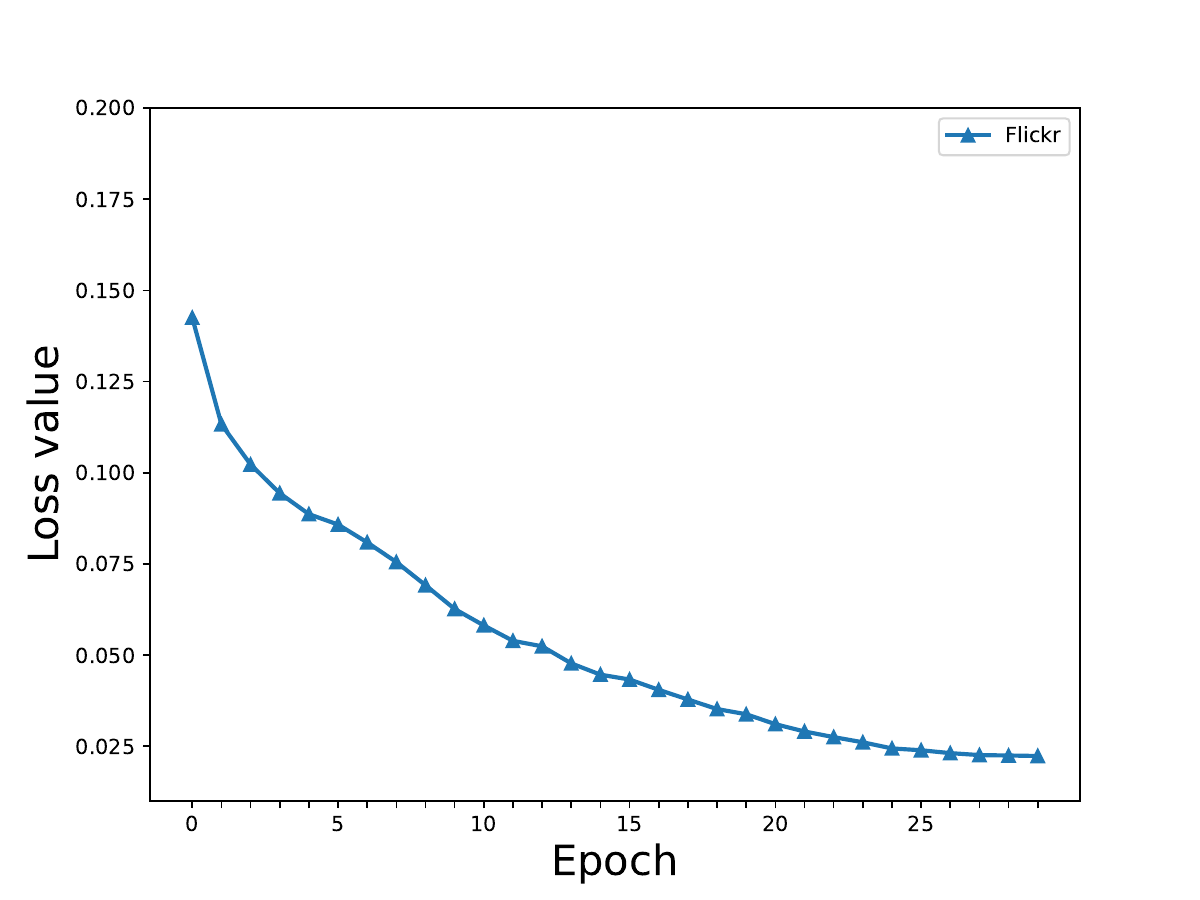}
		\caption{Flickr-LDL}
		\label{fig:flickr_conver}
	\end{subfigure}
	\hfill
	\begin{subfigure}[b]{0.2\textwidth}
		\includegraphics[scale=0.2, trim={0cm 0 1.5cm 0}, clip]{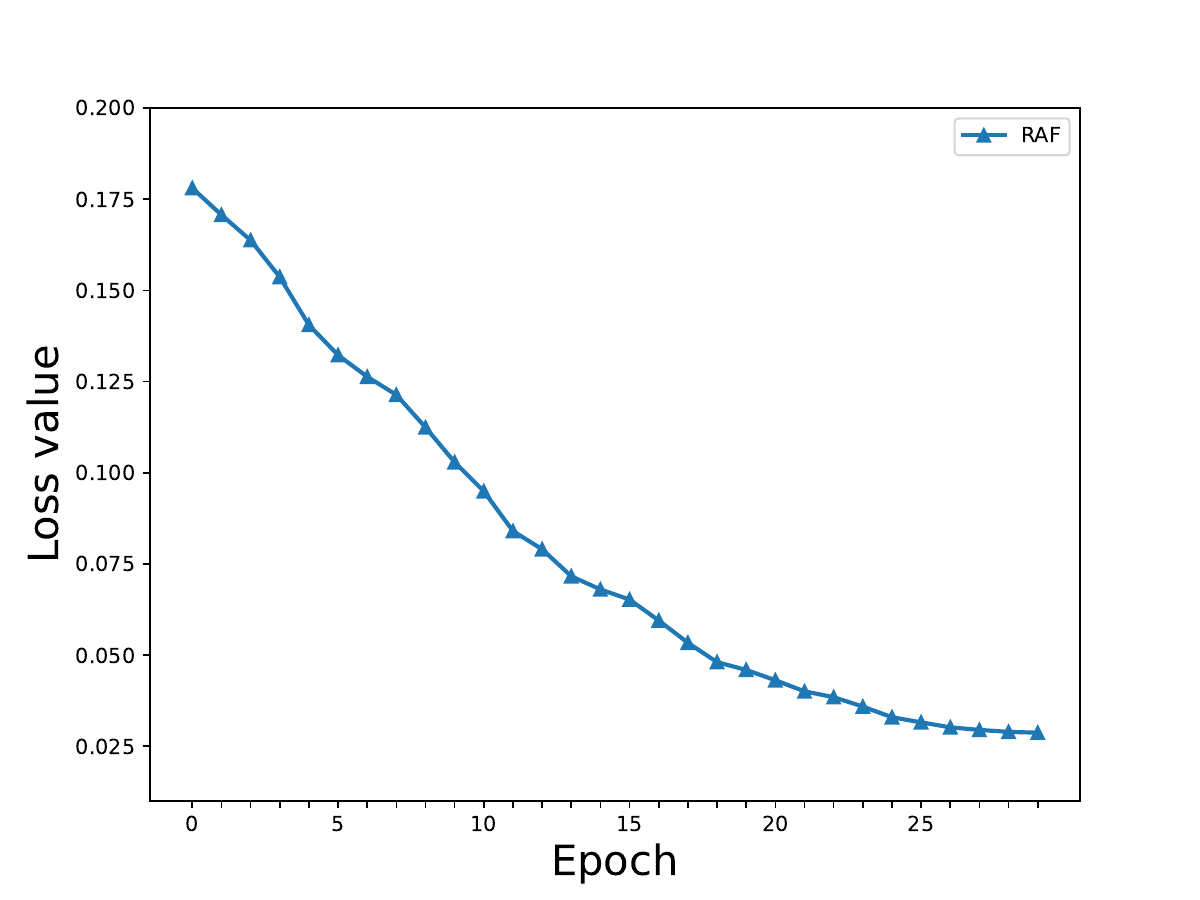}
		\caption{RAF-LDL}
		\label{fig:raf_conver}
	\end{subfigure}
	\hfill
	\begin{subfigure}[b]{0.2\textwidth}
		\includegraphics[scale=0.2, trim={0cm 0 1.5cm 0}, clip]{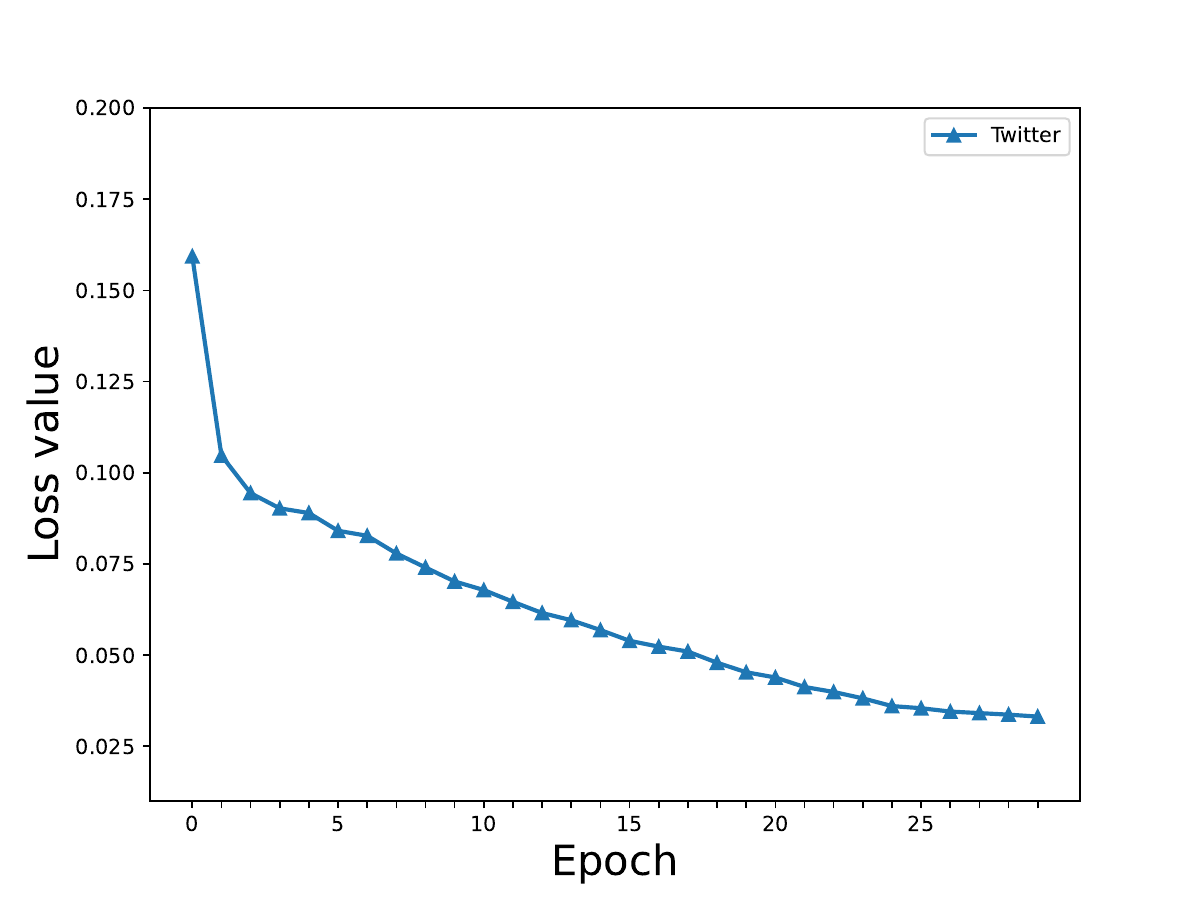}
		\caption{Twitter-LDL}
		\label{fig:twitter_conver}
	\end{subfigure}
	\caption{Convergence on various datasets.}
	\label{shoulianfenxi}
\end{figure*}

\

\begin{figure*}[!ht]
	\centering
	\begin{subfigure}[b]{0.2\linewidth}
		\includegraphics[width=\linewidth]{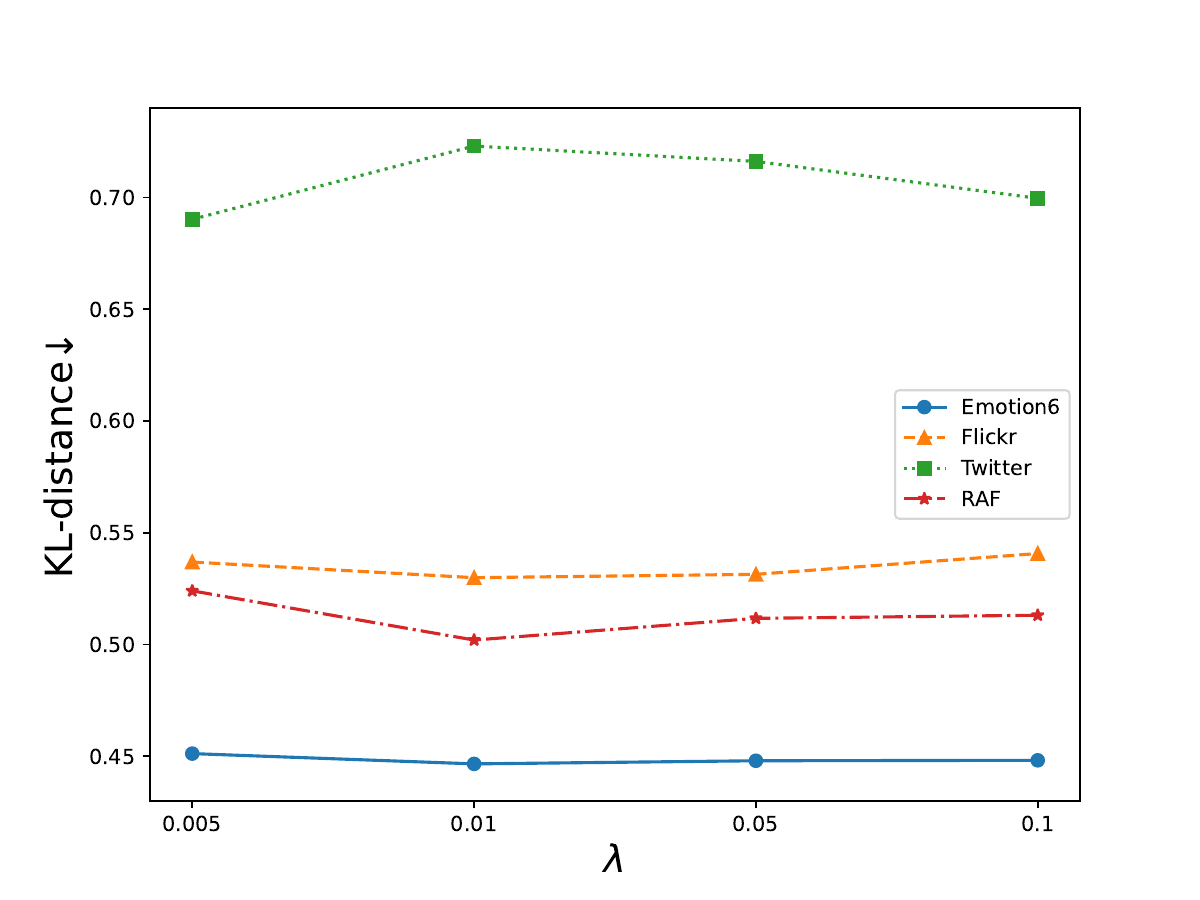}
		\caption{KL with varying $\lambda$}
		\label{fig:kl_lambda}
	\end{subfigure}
	\hfill 
	\begin{subfigure}[b]{0.2\linewidth}
		\includegraphics[width=\linewidth]{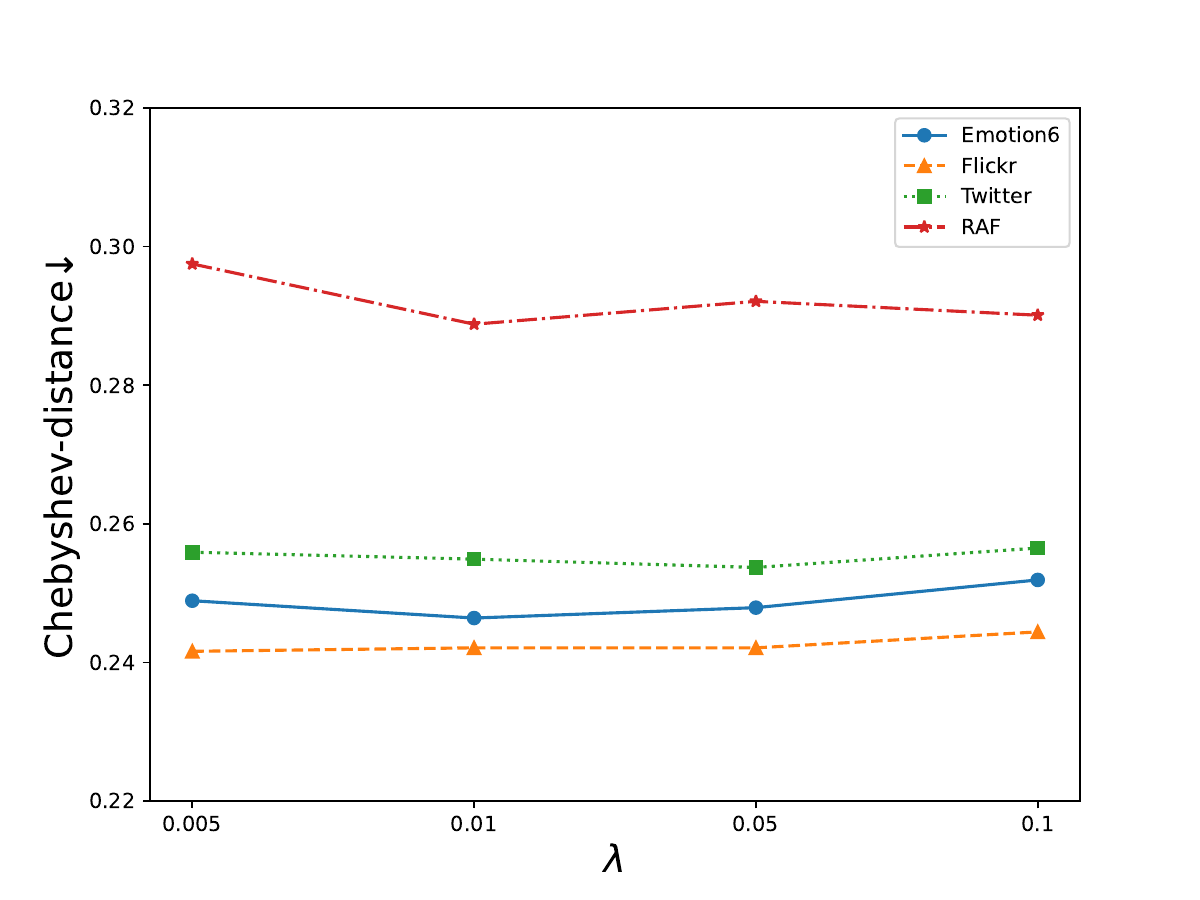}
		\caption{Chebyshev with varying $\lambda$}
		\label{fig:che_lambda}
	\end{subfigure}
	\hfill
	\begin{subfigure}[b]{0.2\linewidth}
		\includegraphics[width=\linewidth]{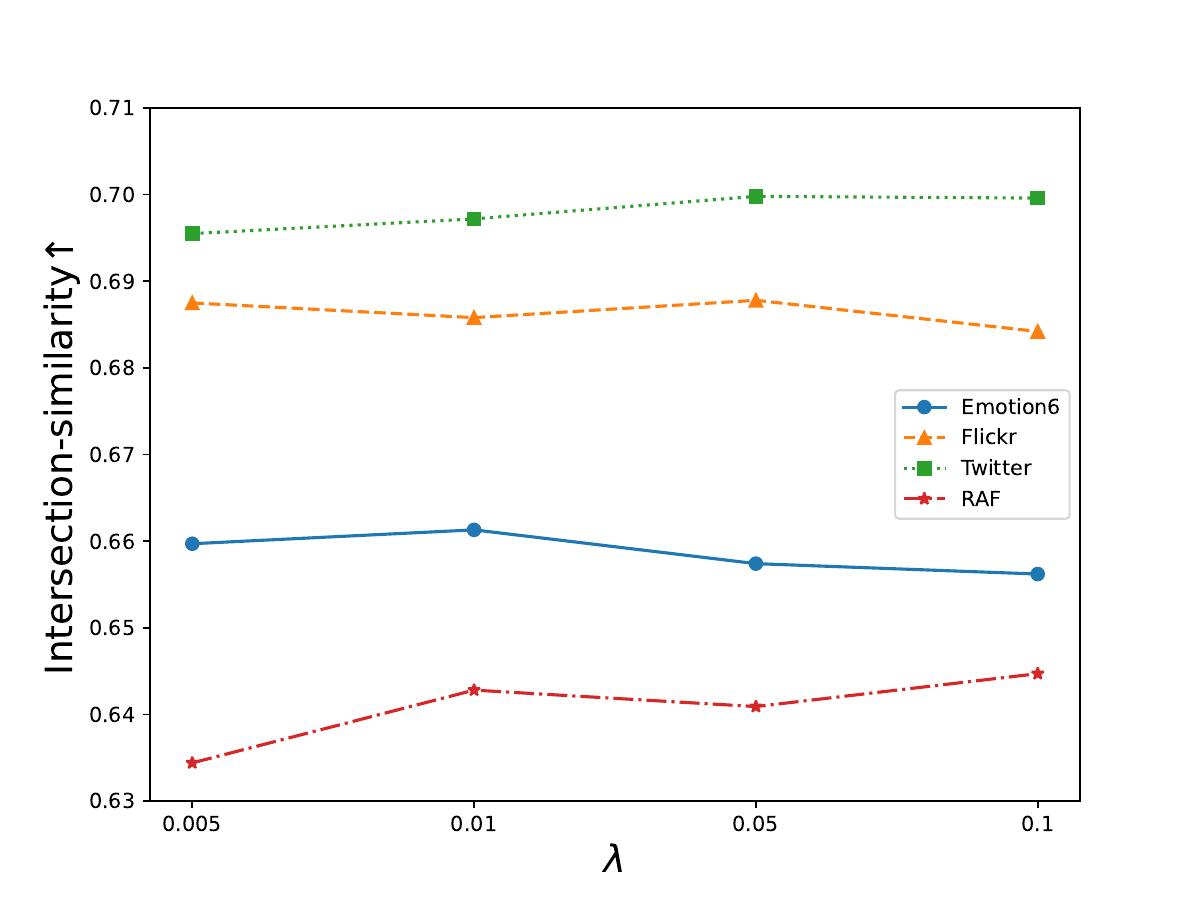}
		\caption{Intersection with varying $\lambda$}
		\label{fig:inter_lambda}
	\end{subfigure}
	\hfill
	\begin{subfigure}[b]{0.2\linewidth}
		\includegraphics[width=\linewidth]{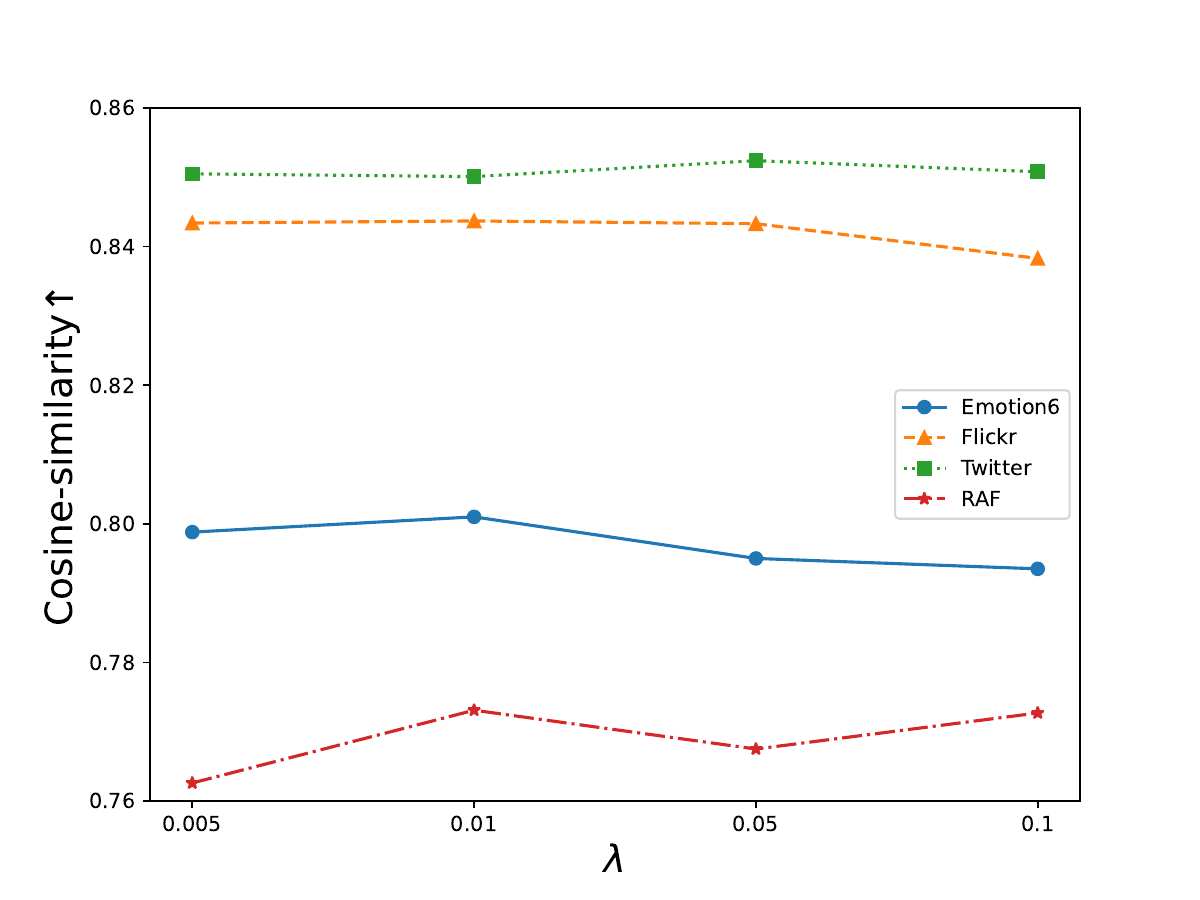}
		\caption{Cosine with varying $\lambda$}
		\label{fig:cosine_lambda}
	\end{subfigure}
	\caption{Convergence on (a) Emotion6, (b) Flickr-LDL, (c) RAF-LDL and (d) Twitter-LDL.}
	\label{canshufenxi}
\end{figure*}

\subsubsection{Ablation Study}
Our ablation study dissected the contributions of the PRR loss and unsupervised consistency loss to RankMatch's effectiveness.  Initially, the model was pre-trained using only 10\% of the labeled data to establish a baseline performance. The results of this stage demonstrated the capability of the model to learn from a limited amount of data. Subsequently, the ranking loss was incorporated into the training process, utilizing the same 10\% of labeled data.  In the final stage, the model was augmented with the unsupervised consistency loss, applied to the unlabeled data. This step built upon the pre-trained model with the supervised ranking loss.  Ablation experiment results are shown in Table. \ref{xiaorongshiyan}. From this, we can draw the following conclusions: 

(a) The PRR loss resulted in a significant improvement over the baseline, emphasizing the component's effectiveness in enhancing the model's discriminative ability, even with limited labeled data. This enhancement accentuates the importance of the ranking mechanism in capturing the complex relationships between labels within a semi-supervised framework. 

(b) The unsupervised consistency loss led to further gains, as demonstrated by the substantial increase in performance presented in the third row of the ablation table. This indicates that the unsupervised loss plays a vital role in harnessing the unlabeled data and refining the model's predictions.

\subsubsection{Convergence Analysis}
In this section, we conduct a convergence analysis of the RankMatch algorithm. The experimental results are shown in Fig. \ref{shoulianfenxi}. The following conclusions are drawn from the observed loss trajectories during training epochs:
For Emotion6, a precipitous decline in loss values during initial epochs underscores the model's rapid learning rate, which then transitions into a stable phase, indicating a quick adjustment to the dataset specifics.
Flickr-LDL mirrors the quick learning pattern of Emotion6 but stabilizes sooner, hinting at an even faster rate of convergence, which could be a characteristic of the dataset or an indication of the model's proficiency in learning.
The loss values for RAF-LDL decrease more gradually, reflecting a consistent and methodical learning process, indicative of the model's steady performance improvement over time.
Twitter-LDL demonstrates a steady downward trend in loss, akin to RAF-LDL, highlighting the model’s consistent learning curve across varying datasets.
Overall, the uniform reduction in loss values across datasets emphasizes the RankMatch algorithm’s capability to optimize the loss function effectively, thereby potentially enhancing the model's prediction accuracy as training advances.

\subsubsection{Parameter Sensitivity Analysis}

In the parameter sensitivity analysis section, we examine the impact of the hyperparameter $\lambda$ on the performance of our semi-supervised label distribution learning algorithm across four different datasets.   The results of the parameter analysis are shown in Fig. \ref{canshufenxi}. From Fig. \ref{canshufenxi},  we can draw the following conclusions: (a)For the KL divergence, the performance remains relatively stable across different values of $\lambda$ on all datasets.  (b)The Chebyshev distance metric shows a mild upward trend as $\lambda$ increases for the Emotion6 and RAF-LDL datasets.  However, this trend is not observed for the Flickr-LDL and Twitter-LDL datasets. (c)Intersection similarity presents a clear trend where performance improves with increasing $\lambda$ on the Emotion6 dataset, but remains relatively stable on the RAF-LDL and Twitter-LDL datasets. The Flickr-LDL dataset shows a slight decline in performance as $\lambda$ increases. (d)The Cosine similarity metric demonstrates robustness to changes in $\lambda$  across all datasets. 

The parameter $\lambda$ has a variable impact on different metrics and datasets. In cases where the performance is affected, the changes are generally gradual and not abrupt, indicating that the RankMatch algorithm is not highly sensitive to this parameter.   This implying that the method is robust to parameter changes and can maintain consistent performance without requiring fine-tuned parameter settings.

\begin{table}[]\centering\scriptsize
	\setlength{\tabcolsep}{5pt}
	\renewcommand{\arraystretch}{1.2}
	\begin{tabular}{ccccccc}
		\hline
		emotion6 & Can $\downarrow$ & Che $\downarrow$ & Clark $\downarrow$ & Cos $\uparrow$ & Inter$\uparrow$ & KL $\downarrow$ \\ \hline
		$\lambda$ =0.01  t = 0.1 & 3.445 & 0.2419 & 1.537 & 0.8037 & 0.6658 & 0.4319 \\
		$\lambda$ =0.01 t = 0.2 & 3.446 & 0.2485 & 1.542 & 0.7977 & 0.6628 & 0.4491 \\
		$\lambda$ =0.01 t = 0.3 & 3.484 & 0.2506 & 1.55 & 0.7933 & 0.6554 & 0.4519 \\
		$\lambda$ =0.01  t = 0.4 & 3.42 & 0.2447 & 1.535 & 0.8067 & 0.6631 & 0.4302 \\ \hline
		RAF-LDL & Can $\downarrow$ & Che $\downarrow$ & Clark $\downarrow$ & Cos $\uparrow$ & Intern $\uparrow$ & KL $\downarrow$ \\ \hline
		$\lambda$ =0.01  t = 0.1 & 3.086 & 0.2954 & 1.467 & 0.7628 & 0.6351 & 0.517 \\
		$\lambda$ =0.01 t = 0.2 & 3.053 & 0.2897 & 1.457 & 0.7726 & 0.6428 & 0.4954 \\
		$\lambda$ =0.01  t = 0.3 & 3.081 & 0.2956 & 1.469 & 0.7643 & 0.6382 & 0.5114 \\
		$\lambda$ =0.01  t = 0.4 & 3.103 & 0.2982 & 1.476 & 0.7612 & 0.6345 & 0.5226 \\ \hline
	\end{tabular}
	\caption{ Impact of Threshold t on the Performance of the RankMatch.}
	\label{yuzhit}
\end{table}

\subsubsection{Impact of Threshold t  on Experimental Results}

The influence of the threshold tin the Pairwise Relevance Ranking loss is critical in determining the sensitivity of the RankMatch algorithm to label ranking discrepancies. Our experiments, as detailed in Table 4, investigate this impact across various datasets and metrics. We observe that the performance metrics on Canberra and Chebyshev distances maintain a relative stability across different tvalues, suggesting that the PRR loss is robust to the threshold variations. The Clark and Cosine distances exhibit a trend where a larger t marginally improves model performance, hinting at the algorithm's enhanced ability to discern more significant label relationships.

Particularly noteworthy is the trend in intersection similarity and KL divergence metrics, especially for the Emotion6 dataset, where a higher threshold t correlates with improved alignment to ground truth labels. This suggests that a careful calibration of t can leverage the RankMatch algorithm's strength in datasets with complex label semantics. The results emphasize the need for a balanced threshold that can effectively distinguish meaningful label importance differences, especially under the constraints of limited labeled data in a semi-supervised learning setting.
\section{Conclusion}
RankMatch advances SSLDL by efficiently combining limited labeled data with a larger volume of unlabeled examples, thus reducing the dependency on extensive manual annotations. It incorporates an averaging strategy inspired by ensemble learning and a pairwise relevance ranking loss, which together enhance prediction stability and model robustness. Our experimental validations across multiple datasets establish RankMatch as a superior method, validating its effectiveness in SSLDL applications.

\nocite{langley00}

\bibliography{example_paper}
\bibliographystyle{icml2022}

\newpage
\appendix
\onecolumn

\end{document}